\documentclass[11pt,a4paper]{article}
\usepackage[hyperref]{acl2019}
\usepackage{times}
\usepackage{latexsym}
\usepackage{breqn}
\usepackage{graphicx}

\usepackage{url}

\usepackage{color}
\newenvironment{graytext}{\color{gray}}{\ignorespacesafterend}

\aclfinalcopy 

\title{Stylized Text Generation Using Wasserstein Autoencoders with a Mixture of Gaussian Prior}

\author{Amirpasha Ghabussi\textsuperscript{1}, Lili Mou\textsuperscript{2}, Olga Vechtomova\textsuperscript{1} \\
\tt{\textsuperscript{1}University of Waterloo, \textsuperscript{2}University of Alberta} \\
\tt{\{aghabuss,ovechtom\}@uwaterloo.ca} \\
\tt{doublepower.mou@gmail.com} \\
}

\date{}

\begin{document}
\maketitle

\begin{abstract}
Wasserstein autoencoders are effective for text generation. They do not however provide any control over the style and topic of the generated sentences if the dataset has multiple classes and includes different topics. In this work, we present a semi-supervised approach for generating stylized sentences. Our model is trained on a multi-class dataset and learns the latent representation of the sentences using a mixture of Gaussian prior without any adversarial losses. This allows us to generate sentences in the style of a specified class or multiple classes by sampling from their corresponding prior distributions. Moreover, we can train our model on relatively small datasets and learn the latent representation of a specified class by adding external data with other styles/classes to our dataset. While a simple WAE or VAE cannot generate diverse sentences in this case, generated sentences with our approach are diverse, fluent, and preserve the style and the content of the desired classes.
\end{abstract}

\section{Introduction}
Probabilistic text generation is an important application of Natural Language Processing (NLP). Variational autoencoder \citep{vae} is a common and important method for sentence generation. VAE imposes a prior distribution on the latent space which is typically set to standard normal. It regularizes the latent space by Kullback-Leibler (KL) divergence \citep{kullback1951information} while reconstructing a data sample. This is equivalent to maximizing the variational lower bound of the likelihood of data. VAE is very difficult to train due to the issue of KL collapse. This can be resolved by adding word dropout or KL annealing to the training process \citep{bowman2015generating}. Another approach to text generation is Generative Adversarial Networks (GAN) \citep{gan}. However, GAN loss is not differentiable and they have difficulties generating discrete sequences \citep{huszar2015not}, therefore VAE seems more appropriate for sentence generation.

Wasserstein autoencoders (WAE) \citep{wae} adjust the aforementioned problems. They regularize the latent space by pushing the aggregated posterior to the prior. This can be achieved by comparing empirical samples from the prior and the posterior distributions. Since, WAE unlike VAE does not push the latent posterior to be close to the prior based on any given input, this results in a better reconstruction performance. Moreover, WAE is much easier to train since it does not use KL divergence to regularize the latent space.

Regular VAE and WAE both generate a sentence by learning a distribution for the latent space. At the inference time, by sampling from this space, they can generate sentences similar to the distribution of the dataset they have been trained on. When the dataset has one class or a topic, this produces satisfactory results. Yet, since they use a standard normal distribution as their prior, they tend to over-regularize the latent space in cases where the dataset consists of multiple classes with different styles or topics. This can be a major drawback of using VAE or WAE for style-specific text generation.

To solve this problem, we propose a WAE with a Gaussian Mixture Prior (GMP) with the number of mixtures set to the number of classes in the dataset. This allows us to generate samples with the style of a specified class by only sampling from the GMP corresponding to this class. Moreover, since we share the same encoder and decoder over all of the classes, we can generate more diverse sentences by training our model on relatively small datasets. Lastly, this allows us to also generate sentences with a mixture of styles by using a weighted average of the latent vectors sampled from multiple Gaussian distributions.

In addition to over regularizing the latent space, most neural networks depend on big datasets and perform poorly when trained on small datasets. However, achieving good results using small data is an important real-world challenge, and in most cases it is harder than solving a big data challenge. With our proposed approach we show that we can train our model on small number of data samples by adding data points from different topics to our data. Our experiments show that this will have very small effect on the style of the generated sentences.

To summarize, our main contributions are:
\begin{itemize}
\item Supervised multi-class sentence generation while preserving the content and style of specified classes
\item Diverse sentence generation on relatively small datasets
\end{itemize}

To evaluate our approach we conduct several experiments. We use the Multi-genre Natural Language Inference (MNLI) dataset \citep{N18-1101} to run all of our experiments. We perform style-conditioned and style-interpolated sentence generation. Our model produces the most diverse sentences among previous works. Moreover we illustrate how our model can outperform others in fluency, diversity, and style accuracy by being trained on a small portion of the dataset.

\section{Related Work}
In natural language processing there is no unique definition of style. Different authors choose a variety of text characteristics as style. Sentiment, formality, genre, and authorship are common choices for representing the style of a sentence \citep{hu2017toward, shen2017style, fu2018style, john2018disentangled}. There are different approaches to style transfer, stylized generation and style-specific topic modeling.








One approach to stylized text generation is using style-specific embeddings for sentence generation. \citet{vechtomova2018generating} use author-specific embeddings to generate stylized poetry, using multi-modal training data. By pretraining the embeddings using a CNN classifier they are able to generate creative data samples. \citet{fu2018style} propose two different approaches for style transfer: style-specific embeddings and style-specific decoder. By applying adversarial losses during training, they encourage the encoder to only include the content of the sentence in the latent space. They use sentiment as the style of a sentence.

Other works focus on learning separate latent representations of style and content for style transfer or stylized generation. \citet{gao2019structuring} use a structured latent space to generate stylized dialogue responses. Their model uses a sequence-to-sequence module and an autoencoder with a shared decoder. \citet{john2018disentangled} propose another approach and apply an adversarial loss to separate style from content. This approach is designed for style transfer, but it can be conditioned on a desired style and used for stylized generation as well.

Mixture of Gaussian prior was previously used for image clustering \citep{ben2018gaussian}. However, using mixture of gaussian for text generation is different from previous works both in terms of the training objective and the model structure. There are different approaches to generate stylized sentences or style transfer. Previous work used Gaussian mixture models as the prior distribution for several NLP tasks. \citet{shen2019mixture} uses Gaussian mixtures for machine translation. \citet{gu2018dialogwae} uses an autoencoder network with a GMP to learn the latent representation of sentence-level data points and jointly trains a GAN to generate and discriminate in the same space. They use the Wasserstein distance to model dialogue responses. \citet{wang2019topic} use an unsupervised approach using a VAE with Gaussian mixture prior for topic modeling. They apply a training penalty to push the Gaussian distributions further apart in the latent space. However, their choice of bag of words for data point representation, does not allow them to generate coherent sentences. Moreover, mixing new data points with their dataset of choice might completely change the topics of their model.

Our work is different from the previous works in that we use a supervised approach with a GMM as our prior distribution and use labeled data for training. Moreover, we refer to a specific topic/class as the style of a sentence similar to \citep{wang2019topic}, but we propose a supervised approach using Wasserstein distance. This allows us to have more control over the specific styles our model will learn. Moreover, it allows us to mix these styles at the inference time. Moreover, we do not apply any penalty to push the Gaussian distributions further away in the latent space and this makes our model easy to train. Finally, we can expand our dataset and add new training samples to help our encoder to effectively learn the latent representation of our desired classes, and help our decoder to generate much more diverse sentences.

\section{Approach}
In this section we describe our approach in detail. We use a stochastic WAE with MMD penalty with a sequence to sequence neural network  \citep{sutskever2014sequence}. Using a Gaussian mixture distribution for our prior we are able to generate single and multi-style conditioned sentences at the inference time. We further explain our training process and the details of our model in this section.

\subsection{Autoencoder}
Autoencoders \citep{baldi2012autoencoders} encode an input into a latent representation, from which they reconstruct the input again. Usually, the input has a much higher dimension than its corresponding latent representation. However, in some cases, such as noise reduction and text enhancement, the latent representation can have higher dimensions \citep{lu2013speech}. Another important application of autoencoders is style transfer \citep{shen2017style}. 

Depending on the task, there are multiple design options for the encoder and the decoder networks of an autoencoder, which are chosen based on the input structure. In natural language processing, a common choice for these networks is a feed-forward neural network when the input format is bag-of-words (BOW) \citep{wallach2006topic}. Another common choice for input sequences are Recurrent Neural Networks (RNN). In this work, we use Gated Recurrent Units (GRU) \citep{choi2016doctor} as our choice of the RNN cell for our encoder and decoder.

Given that at time step $t$ the decoder predicts the next token to be $x_t$ the training loss of the autoencoder $J_{AE}$ is definded as:
\begin{dmath}
J_{AE} = \sum\limits_{i = 1}^N \sum\limits_{t = 1}^T - log(p(x_t | h, x_1, x_2, ..., x_{t-1}))
\end{dmath}
where $h$ is the latent vector representation, $N$ is the number of training samples, and $T$ is the total number of decoding steps.

\subsection{Wasserstein Autoencoder}
One approach to regularize the posterior is to impose a constraint that the aggregated posterior of $h$ should be similar to its prior \citep{wae}. This constraint can be relaxed by penalizing the Wasserstein distance between $q(h)$ and $p(h)$. This can be computed as the Maximum Mean Discrepancy (MMD) between $Q(h)$ and $P(h)$:

\begin{dmath}
MMD = \left\Vert \int k(h,.) dP(h) - \int k(h,.) dQ(h)\right\Vert_{H_k}
\end{dmath}

Where $H_k$ is the reproducing kernel Hilbert space defined by kernel $k$. We chose the inverse multi-quadratic kernel $k(x, y) = \frac{C}{C + ||x - y||_2^2}$ in our experiments which is a common choice.

The MMD penalty can be estimated by empirical samples as:
\begin{dmath}
  \widehat{MMD} = \frac{1}{N(N -1)}
\left(\sum\limits_{n \neq m} k(h^{(n)}, h^{(m)}) + \sum\limits_{n \neq m} k(\widetilde{h}^{(n)}, \widetilde{h}^{(m)})\right)
- \frac{1}{N^2} \sum\limits_{n, m} k(h^{(n)}, \widetilde{h}^{(m)}) 
\label{mmd}
\end{dmath}
where $\widetilde{h}^{(n)}$ is a sample from prior $p$ and $h^{(n)}$ is a sample from the aggregated posterior $q$.

\subsection{WAE with Gaussian Mixture Model Prior (GMM-WAE)}
In this work we use a Gaussian mixture model as the chosen distribution for our WAE prior. There are multiple benefits gained from this. First, many datasets are a combination of different styles and classes, therefore, the  model structure should account for this in order to learn a good representation of these datasets. Moreover, separating the latent representation of a group of data samples, allows the model to be trained on completely different data points at the same time and learn multiple latent distributions independently. The final distribution of our latent space follows the Gaussian mixture model distribution:

\begin{dmath}
P(z) = \sum\limits_{i = 1}^N w_i \mathcal{N}(\mu_i, \sigma_i)
\end{dmath}

Where $N$ is the number of mixture distributions, $\Sigma_{i = 1}^N w_i = 1$, and $w_i \geq 0$. If a dataset has $N$ classes with distinct styles, we use the same number of Gaussian distributions for our latent space and encode every sentence to its corresponding latent distribution. Then, the latent vector representation is defined as:

\begin{dmath}
h = \sum\limits_{i = 1}^N w_i \times h_i
\label{latent-vector}
\end{dmath}

Where $h_i$ denotes the sampled vector from the $i_{th}$ Gaussian mixture distribution $h_i \sim \mathcal{N}(\mu_i, \sigma_i)$ and $w_i$ is its corresponding weight. 

\subsection{Training}
At the training time, each input sequence $(x_1^i, x_2^i, ..., x_n^i)$ is mapped to its corresponding mean and variance vectors. We simultaneously learn multiple priors by pushing the encoded mean and variance vectors to their corresponding prior mean $(\mu_i)$ and variance $(\sigma_i)$ vectors. Since we use a stochastic WAE, we then sample from a normal distribution with the same encoded mean and a variance of $1$. We use KL-divergence to regularize the stochastic part of our model and produce more diverse sentences based on the following objective:

\begin{dmath}
J_{KL} = \sum \limits_{i = 0}^N KL \left( \mathcal{N}(\mu_{post}, diag(\sigma_{post})^2) || \mathcal{N}(\mu_{post}, \textbf{I}) \right)
\end{dmath}

To regularize our latent space and learn the prior distribution, we use the MMD penalty following Equation \ref{mmd}. The final training loss is the weighted sum of the KL loss, MMD loss, and reconstruction loss. Hence, it can be written as:

\begin{dmath}
J_{WAE} = J_{AE} + \lambda_{KL} \cdot J_{KL} + \lambda_{MMD} \cdot \sum\limits_{j = 0}^M \times \widehat{MMD_j}
\end{dmath}

Where M is the number of classes in our dataset. This is our training objective at the training phase. Note that the gradient will only back-propagate through the Gaussian distribution corresponding to the batch class.

\begin{figure}
  \includegraphics[width=\linewidth]{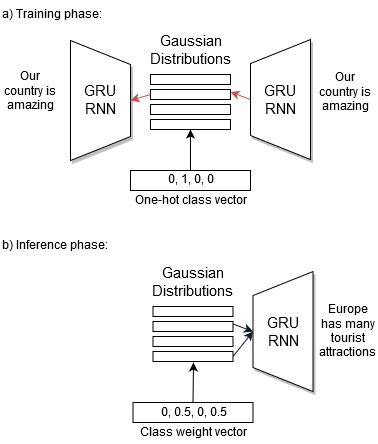}
  \caption{Overview of our approach. The red arrows represent backpropagation through a single class. The black arrows represent the forward pass where the hidden vector is sampled from two classes with equal weights.}
  \label{approach}
\end{figure}

During the training phase, we use mini-batches where the samples are from only one input class. This is a stochastic estimation of the actual gradient descent algorithm. Individual batches are biased towards a certain class, but with multiple batches sampled from all of the classes we estimate the actual training objective. For a sequence with class $i$ we set all other latent weights to zero and $w_i = 0$. This allows us to only back-propagate the reconstruction loss through the $i_{th}$ Gaussian distribution and the MMD penalty will push $\mu_i$ and $\sigma_i$ to the encoded vectors. Moreover, we use recurrent architectures for encoder and decoder and  cross-entropy loss for reconstruction. Figure \ref{approach}a shows an overview of the training process. The one-hot class vector is the training weights for the mixture distributions and the red arrows show the backpropagation through just one of the distributions.

\subsection{Sentence Generation with GMM-WAE}
Text generation with GMM-WAE is slightly different from the training process. By sampling from the latent space we can generate new sentences conditioned  either on a single class, or on multiple classes. To generate a sentence, we first have to sample from the latent space and produce the latent vector $h$ following Equation \ref{latent-vector}. Then we simply feed this vector to the recurrent decoder as its initial state, and  append it to the input of every time step. We use the standard inference decoder following \citet{wu2016google}. Figure \ref{approach}b shows an overview of the inference process. The classes contributing to the style of the final sentence are the weights with non-zero values in the class vector.

\begin{table*}[t!]
\centering
\begin{tabular}{llllll}
   & D-1$\uparrow$ & D-2$\uparrow$ & Entropy$\uparrow$ & (PPL)$\downarrow$ & Classification$\uparrow$\\
  \hline
  \multicolumn{6}{c}{Trained on MNLI}\\
  \hline
    Lyrics VAE&
    0.034 &
    0.070 &
    4.153 &
    112.3 &
    85.2\\
    Dinentangled VAE &
    0.027 &
    0.117 &
    4.853 &
    \textbf{73.81} &
    77.8 \\
    separate WAE &
    0.052 &
    0.214 &
    5.416 &
    95.2 &
    \begin{graytext}97.9\end{graytext}\\
    GMM WAE (ours) &
    \textbf{0.061} &
    \textbf{0.392} &
    \textbf{5.673} &
    98.4 &
    \textbf{85.8}\\
  \hline
  \multicolumn{6}{c}{Trained on 40960 samples from MNLI}\\
  \hline
    Lyrics VAE(s)&
    0.021 &
    0.055 &
    4.001 &
    112.3 &
    82.4\\
    Disentangled VAE(s)&
    0.021 &
    0.096 &
    4.392 &
    \textbf{101.1} &
    75.3 \\
    separate WAE(s)&
    0.017 &
    0.193 &
    4.920 &
    105.7 &
    \begin{graytext}92.2\end{graytext}\\
    GMM WAE (ours)(s)&
    \textbf{0.049} &
    \textbf{0.284} &
    \textbf{5.201} &
    \textbf{101.2} &
    \textbf{84.0}\\
\end{tabular}
\caption{Comparison between our work and others. Note that the classification accuracy for separate WAE models is not a valid measure since each model is only trained on a single class. Hence the accuracy should be 100\% in theory. Models with (s) in their names, are trained on a small subset of MNLI with 40960 raining samples for four classes}
\label{table:table1}
\end{table*}

\textbf{Style-Conditioned Sentence Generation}: In this setup, we generate sentences conditioned on a single style. This process is similar to the training approach. We set all $w_i$ to zero except for the weight of the class, corresponding to the desired (target) style of the generated sentence. Hence the latent vector is sampled from $P(z) = \mathcal{N}(\mu_k, \sigma_k)$ where $k$ is the target style the generated sentence is conditioned on. The sampled latent vector will only include features from the target class. This is very similar to what we perform at training time, so the results are expected to be very good.
    
\textbf{Style-Interpolated Sentence Generation}: In this second setup, generated sentences are conditioned on an interpolation between two latent vector samples. For generating a sentence with more than one style, we simply interpolate between two samples from the mixture Gaussian distributions. We set two $w_i$s to non-zero values while satisfying the condition that $\Sigma_i w_i = 1$. This is an equivalent of a weighted average between the latent vector samples. By changing the value of the weights for each distribution, we can control the contribution of each style in the final generated sentence. For the sake of our experiments two of the $w_i$ weights are set to $0.5$ and the rest are zero. This means the latent vector is sampled from $P(z) = \frac{1}{2} (\mathcal{N}(\mu_{k1}, \sigma_{k1}) + \mathcal{N}(\mu_{k2}, \sigma_{k2}))$ with $k1$ and $k2$ being the desired classes.

\section{Experiments}
To evaluate our approach we use the MNLI. We use a sequence to sequence \citep{sutskever2014sequence} setup with maximum sequence length of 30. Our vocabulary size is 30,000 and our latent space has 100 dimensions for every Gaussian prior. During the training process, we append the encoded latent vector to every step of the decoder. For inference, we use the generated token at each time step as the input to the next RNN time step and append the sampled latent vector to all decoder time steps, similar to the training process. We compare our results with the work of \cite{vechtomova2018generating}, and other baselines.

MNLI consists of 433k crowd-sourced sentences from five different genres: Slate, Telephone, Government, Fiction, and Travel. We ignore the Slate genre in our experiments since the sentences in this genre cover a diverse set of topics and it confuses our model. We run two experiments on MNLI to evaluate the performance of our model. Our first experiment uses all of the sentences available and compares the stylized generation performance of our approach with previous work and baselines. For our second experiment, we use a subset of 10240 sentences from the four classes mentioned above. We generate samples using separate WAEs trained on sentences from individual classes and using a WAE with GM prior trained on all 40960 sentences. We describe the metrics used for our comparison in the results section.

\begin{table*}[t!]
\begin{center}
\begin{tabular}{|l|}
  \hline
  \textbf{Fiction}\\
  \hline
  kramenin? he drew the question the last time that 's happened?\\
  man , apparently you do n't think that the doctor 's always alone.\\
  \hline
  \textbf{Government}\\
  \hline
  i provide guidance in determining the requirements of state agencies also used the additional databases.\\
  there are no success of delivery in california , reducing in pm concentrations.\\
  \hline
  \textbf{Travel}\\
  \hline
  hong kong is now a fascinating fifth-century , walled architecture.\\
  the greatest can be sensed in dublin and its surrounding farmlands , a full of historic buildings.\\
  \hline
  \textbf{Telephone}\\
  \hline
  uh-huh yeah i guess you ca n't have our problem.\\
  so they had to talk about it, um oh absolutely.\\
  \hline
  \textbf{Government + Travel}\\
  \hline
  so we 've talked to our children to pursue little observation from the standpoint that we 're split in.\\
  one provides opportunities for bargaining delivery system to link between gagas and research is helpful.\\
  \hline
  \textbf{Government + Telephone}\\
  \hline
  8 time, i mean you have UNK UNK outside the new government.\\
  when the general requires a current protections of federal acquisition , he said an organization \\ had adoptedeach retiree.\\
  \hline
  \textbf{Travel + Telephone}\\
  \hline
  given the book now i know like that capital or UNK egypt who came to conquer.\\
  that i guess the remaining states that now is a more easily protected\\
\hline
\end{tabular}
\end{center}
\caption{Sentences generated by our GMP WAE model.}
\end{table*}

Our model works best for generating diverse sentences and outperforms other models in most of the evaluation metrics. When the dataset is relatively small, a WAE or VAE do not capture enough features to generate diverse and fluent sentences. We use a WAE with GMP and train our model using ten percent of the data in MNLI and compare its performance with other models. Our model outperforms all of the other models in this case.

\subsection{Evaluataion}
In this section we discuss different metrics we used to evaluate the performance of our model and compare it with baselines, and previous work. We use Jansen-Shannon Divergence to evaluate style-interpolated sentence generation classification accuracy. Also we use multiple measures for sentence diversity and finally we use perplexity to validate the coherency and fluency of the generated sentences.

\textbf{Jensen-Shannon Divergence:} Jensen-Shannon Divergence (JSD) \cite{lin1991divergence} is our metric of choice to evaluate our multi-class sentence inference accuracy. We sample from two of the Gaussian distributions and average the sampled vectors with equal weights. Then, we feed this vector to the decoder. To determine the class of the inferred sentence, we use our pre-trained classifier. In the ideal situation, the output of the last layer of the classifier should be 0 for all of the classes and $0.5$ for the two sampled classes. JSD quantifies the difference between the sampling probability distribution and the classification distribution of the sampled sentences.



\begin{table*}[t!]
\centering
\begin{tabular}{llllll}
   & Gov & Fic & Trav & Tel & JSD\\
  \hline
Gov &
\textbf{70.27} &
03.87 &
19.98 &
05.88 &
0.116\\
Fic &
00.31 &
\textbf{96.58} &
03.03 &
00.08 &
0.012\\
Trav &
01.76 &
09.94 &
\textbf{87.68} &
00.63 &
0.045\\
Tel &
05.25 &
03.28 &
02.46 &
\textbf{89.00} &
0.040\\

\end{tabular}
\caption{Classification acuracy and JSD values for GMM-WAE Style-conditioned sentence generation using 40960 training samples from MNLI.
  }
\end{table*}

\begin{table*}[t!]
\centering
\begin{tabular}{llllll}
   & Gov & Fic & Trav & Tel & JSD\\
  \hline
Fic-Gov &
04.84 &
\textbf{41.33}&
\textbf{50.21} &
03.61 &
0.297 \\
Fic-Trav &
01.29 &
\textbf{61.41} &
\textbf{36.56} &
07.42 &
0.037\\
Fic-Tel &
26.30 &
\textbf{38.13} &
\textbf{29.12} &
06.44 &
0.292\\
Gov-Trav &
\textbf{21.76} &
11.05 &
\textbf{64.71} &
02.48 &
0.080\\
Gov-Tel &
\textbf{39.47} &
04.80 &
10.00 &
\textbf{45.72} &
0.055\\t
Trav-Tel &
\textbf{41.21} &
04.57 &
15.78 &
\textbf{38.44} &
0.209\\
\end{tabular}

\caption{Single and Multi-class sentence generation for 40960 sentences from MNLI using WAE with GM Prior. Each row represents the classifier's confidence for sampled Sentences from two classes. The sampling weight is equal to 50\% for both classes. In all of the cases the classifier classifies the single-class samples correctly and for multi-class samples it always finds at least one of the correct classes. However, Half of the times it correctly finds the other class as well. This table is good for understanding and comparing how a classification and sampling distribution translates into JSD values.
  }
\end{table*}

\textbf{Style Accuracy:} We follow the approach of the previous work \cite{hu2017toward}, \cite{shen2017style}, \cite{fu2018style}, \cite{john2018disentangled} and separately train a convolutional neural network (CNN) to classify sentences \cite{kim2014convolutional} based on their classes. We use this classifier to classify sentences generated with our approach and compare our results with a separate WAE trained only on a specified class of a dataset. We also ran our experiments with separate VAEs and the results are very close to separate WAEs. The classification accuracy of the classifier over the original MNLI dataset is 98\%. Table \ref{table:table1} compares the classification results. Using the WAE with a GM prior lowers the accuracy of the classifier. This is expected because we use a shared decoder over multiple class distributions, but the classifier still is easily able to identify different classes.

\textbf{Perplexity: }
We use the Kneser-Ney language model \cite{kneser1995improved} to evaluate the fluency of our sampled sentences. We measure the empirical distribution of trigrams in a corpus, and compute the log-likelihood of a sentence. We train the language model on the original dataset and evaluate the fluency of our sampled sentences. \ref{table:table1} provides the fluency results. Our model achieves acceptable results compared to separate WAEs trained on only one class of sentences. This is again because we are sharing the decoder for all of the GM priors and thus, the decoder sometimes uses rare words to generate a sentence for a specified class. These rare words might be common words in another class. Since the language model has not seen such a combination of words, it evaluates these sentences with a lower accuracy. Note that smaller values correspond to more fluent sentences.

\textbf{Diversity: }
We use distinct diversity metrics by computing the percentage of distinct unigrams and bigrams following the work of \cite{li2015diversity} and \cite{bahuleyan2017variational}. Our model outperforms all of the baselines and previous models in terms of sentence diversity. This is because the decoder learns on a more diverse set of sentences when it is trained over multiple classes. The diversity results are provided in \ref{table:table1}. 
For the question generation task, since the dataset is very small, neither of the models are successful at generating diverse sentences and they tend to generate the same set of question over and over. However, WAE with GM prior generates twice more diverse sentences compared to separate WAEs.

\section{Conclusion}
Compared to WAE and VAE, WAE with GMP provides control over the style of generated samples. Moreover, it generates fluent and diverse sentences while it is capable of generating sentences with a mixture of styles. Additionally, since the GMP is powerful to capture the latent representation of the dataset, it is possible to add more data samples with other classes to small datasets and learn enough features to generate diverse samples with a desired style/class. The VAE and WAE are not capable of learning the representation of a dataset, nor can they learn a good language model when the dataset has few training samples.

\newpage

\bibliography{references.bib}

\begin{thebibliography}{28}
\expandafter\ifx\csname natexlab\endcsname\relax\def\natexlab#1{#1}\fi

\bibitem[{Bahuleyan et~al.(2017)Bahuleyan, Mou, Vechtomova, and
  Poupart}]{bahuleyan2017variational}
Hareesh Bahuleyan, Lili Mou, Olga Vechtomova, and Pascal Poupart. 2017.
\newblock Variational attention for sequence-to-sequence models.
\newblock \emph{arXiv preprint arXiv:1712.08207}.

\bibitem[{Baldi(2012)}]{baldi2012autoencoders}
Pierre Baldi. 2012.
\newblock Autoencoders, unsupervised learning, and deep architectures.
\newblock In \emph{Proceedings of ICML workshop on unsupervised and transfer
  learning}, pages 37--49.

\bibitem[{Ben-Yosef and Weinshall(2018)}]{ben2018gaussian}
Matan Ben-Yosef and Daphna Weinshall. 2018.
\newblock Gaussian mixture generative adversarial networks for diverse
  datasets, and the unsupervised clustering of images.
\newblock \emph{arXiv preprint arXiv:1808.10356}.

\bibitem[{Bowman et~al.(2015)Bowman, Vilnis, Vinyals, Dai, Jozefowicz, and
  Bengio}]{bowman2015generating}
Samuel~R Bowman, Luke Vilnis, Oriol Vinyals, Andrew~M Dai, Rafal Jozefowicz,
  and Samy Bengio. 2015.
\newblock Generating sentences from a continuous space.
\newblock \emph{arXiv preprint arXiv:1511.06349}.

\bibitem[{Choi et~al.(2016)Choi, Bahadori, Schuetz, Stewart, and
  Sun}]{choi2016doctor}
Edward Choi, Mohammad~Taha Bahadori, Andy Schuetz, Walter~F Stewart, and Jimeng
  Sun. 2016.
\newblock Doctor ai: Predicting clinical events via recurrent neural networks.
\newblock In \emph{Machine Learning for Healthcare Conference}, pages 301--318.

\bibitem[{Fu et~al.(2018)Fu, Tan, Peng, Zhao, and Yan}]{fu2018style}
Zhenxin Fu, Xiaoye Tan, Nanyun Peng, Dongyan Zhao, and Rui Yan. 2018.
\newblock Style transfer in text: Exploration and evaluation.
\newblock In \emph{Thirty-Second AAAI Conference on Artificial Intelligence}.

\bibitem[{Gao et~al.(2019)Gao, Zhang, Lee, Galley, Brockett, Gao, and
  Dolan}]{gao2019structuring}
Xiang Gao, Yizhe Zhang, Sungjin Lee, Michel Galley, Chris Brockett, Jianfeng
  Gao, and Bill Dolan. 2019.
\newblock Structuring latent spaces for stylized response generation.
\newblock \emph{arXiv preprint arXiv:1909.05361}.

\bibitem[{Goodfellow et~al.(2014)Goodfellow, Pouget-Abadie, Mirza, Xu,
  Warde-Farley, Ozair, Courville, and Bengio}]{gan}
Ian Goodfellow, Jean Pouget-Abadie, Mehdi Mirza, Bing Xu, David Warde-Farley,
  Sherjil Ozair, Aaron Courville, and Yoshua Bengio. 2014.
\newblock Generative adversarial nets.
\newblock In \emph{Advances in neural information processing systems}, pages
  2672--2680.

\bibitem[{Gu et~al.(2018)Gu, Cho, Ha, and Kim}]{gu2018dialogwae}
Xiaodong Gu, Kyunghyun Cho, Jung-Woo Ha, and Sunghun Kim. 2018.
\newblock Dialogwae: Multimodal response generation with conditional
  wasserstein auto-encoder.
\newblock \emph{arXiv preprint arXiv:1805.12352}.

\bibitem[{Hu et~al.(2017)Hu, Yang, Liang, Salakhutdinov, and
  Xing}]{hu2017toward}
Zhiting Hu, Zichao Yang, Xiaodan Liang, Ruslan Salakhutdinov, and Eric~P Xing.
  2017.
\newblock Toward controlled generation of text.
\newblock In \emph{Proceedings of the 34th International Conference on Machine
  Learning-Volume 70}, pages 1587--1596. JMLR. org.

\bibitem[{Husz{\'a}r(2015)}]{huszar2015not}
Ferenc Husz{\'a}r. 2015.
\newblock How (not) to train your generative model: Scheduled sampling,
  likelihood, adversary?
\newblock \emph{arXiv preprint arXiv:1511.05101}.

\bibitem[{John et~al.(2018)John, Mou, Bahuleyan, and
  Vechtomova}]{john2018disentangled}
Vineet John, Lili Mou, Hareesh Bahuleyan, and Olga Vechtomova. 2018.
\newblock Disentangled representation learning for text style transfer.
\newblock \emph{arXiv preprint arXiv:1808.04339}.

\bibitem[{Kim(2014)}]{kim2014convolutional}
Yoon Kim. 2014.
\newblock Convolutional neural networks for sentence classification.
\newblock \emph{arXiv preprint arXiv:1408.5882}.

\bibitem[{Kingma and Welling(2013)}]{vae}
Diederik~P Kingma and Max Welling. 2013.
\newblock Auto-encoding variational bayes.
\newblock \emph{arXiv preprint arXiv:1312.6114}.

\bibitem[{Kneser and Ney(1995)}]{kneser1995improved}
Reinhard Kneser and Hermann Ney. 1995.
\newblock Improved backing-off for m-gram language modeling.
\newblock In \emph{1995 International Conference on Acoustics, Speech, and
  Signal Processing}, volume~1, pages 181--184. IEEE.

\bibitem[{Kullback and Leibler(1951)}]{kullback1951information}
Solomon Kullback and Richard~A Leibler. 1951.
\newblock On information and sufficiency.
\newblock \emph{The annals of mathematical statistics}, 22(1):79--86.

\bibitem[{Li et~al.(2015)Li, Galley, Brockett, Gao, and
  Dolan}]{li2015diversity}
Jiwei Li, Michel Galley, Chris Brockett, Jianfeng Gao, and Bill Dolan. 2015.
\newblock A diversity-promoting objective function for neural conversation
  models.
\newblock \emph{arXiv preprint arXiv:1510.03055}.

\bibitem[{Lin(1991)}]{lin1991divergence}
Jianhua Lin. 1991.
\newblock Divergence measures based on the shannon entropy.
\newblock \emph{IEEE Transactions on Information theory}, 37(1):145--151.

\bibitem[{Lu et~al.(2013)Lu, Tsao, Matsuda, and Hori}]{lu2013speech}
Xugang Lu, Yu~Tsao, Shigeki Matsuda, and Chiori Hori. 2013.
\newblock Speech enhancement based on deep denoising autoencoder.
\newblock In \emph{Interspeech}, pages 436--440.

\bibitem[{Shen et~al.(2017)Shen, Lei, Barzilay, and Jaakkola}]{shen2017style}
Tianxiao Shen, Tao Lei, Regina Barzilay, and Tommi Jaakkola. 2017.
\newblock Style transfer from non-parallel text by cross-alignment.
\newblock In \emph{Advances in neural information processing systems}, pages
  6830--6841.

\bibitem[{Shen et~al.(2019)Shen, Ott, Auli, and Ranzato}]{shen2019mixture}
Tianxiao Shen, Myle Ott, Michael Auli, and Marc'Aurelio Ranzato. 2019.
\newblock Mixture models for diverse machine translation: Tricks of the trade.
\newblock \emph{arXiv preprint arXiv:1902.07816}.

\bibitem[{Sutskever et~al.(2014)Sutskever, Vinyals, and
  Le}]{sutskever2014sequence}
Ilya Sutskever, Oriol Vinyals, and Quoc~V Le. 2014.
\newblock Sequence to sequence learning with neural networks.
\newblock In \emph{Advances in neural information processing systems}, pages
  3104--3112.

\bibitem[{Tolstikhin et~al.(2017)Tolstikhin, Bousquet, Gelly, and
  Schoelkopf}]{wae}
Ilya Tolstikhin, Olivier Bousquet, Sylvain Gelly, and Bernhard Schoelkopf.
  2017.
\newblock Wasserstein auto-encoders.
\newblock \emph{arXiv preprint arXiv:1711.01558}.

\bibitem[{Vechtomova et~al.(2018)Vechtomova, Bahuleyan, Ghabussi, and
  John}]{vechtomova2018generating}
Olga Vechtomova, Hareesh Bahuleyan, Amirpasha Ghabussi, and Vineet John. 2018.
\newblock Generating lyrics with variational autoencoder and multi-modal artist
  embeddings.
\newblock \emph{arXiv preprint arXiv:1812.08318}.

\bibitem[{Wallach(2006)}]{wallach2006topic}
Hanna~M Wallach. 2006.
\newblock Topic modeling: beyond bag-of-words.
\newblock In \emph{Proceedings of the 23rd international conference on Machine
  learning}, pages 977--984. ACM.

\bibitem[{Wang et~al.(2019)Wang, Gan, Xu, Zhang, Wang, Shen, Chen, and
  Carin}]{wang2019topic}
Wenlin Wang, Zhe Gan, Hongteng Xu, Ruiyi Zhang, Guoyin Wang, Dinghan Shen,
  Changyou Chen, and Lawrence Carin. 2019.
\newblock Topic-guided variational autoencoders for text generation.
\newblock \emph{arXiv preprint arXiv:1903.07137}.

\bibitem[{Williams et~al.(2018)Williams, Nangia, and Bowman}]{N18-1101}
Adina Williams, Nikita Nangia, and Samuel Bowman. 2018.
\newblock \href {http://aclweb.org/anthology/N18-1101} {A broad-coverage
  challenge corpus for sentence understanding through inference}.
\newblock In \emph{Proceedings of the 2018 Conference of the North American
  Chapter of the Association for Computational Linguistics: Human Language
  Technologies, Volume 1 (Long Papers)}, pages 1112--1122. Association for
  Computational Linguistics.

\bibitem[{Wu et~al.(2016)Wu, Schuster, Chen, Le, Norouzi, Macherey, Krikun,
  Cao, Gao, Macherey et~al.}]{wu2016google}
Yonghui Wu, Mike Schuster, Zhifeng Chen, Quoc~V Le, Mohammad Norouzi, Wolfgang
  Macherey, Maxim Krikun, Yuan Cao, Qin Gao, Klaus Macherey, et~al. 2016.
\newblock Google's neural machine translation system: Bridging the gap between
  human and machine translation.
\newblock \emph{arXiv preprint arXiv:1609.08144}.

\end{thebibliography}
\bibliographystyle{acl_natbib}

\appendix

\end{document}